# Image Analytics for Legal Document Review
## A Transfer Learning Approach

Nathaniel Huber-Fliflet
Data & Technology
Ankura
Washington DC, USA
nathaniel.huber-fliflet@ankura.com

Fusheng Wei
Data & Technology
Ankura
Washington DC, USA
fusheng.wei@ankura.com

Haozhen Zhao
Data & Technology
Ankura
Washington DC, USA
haozhen.zhao@ankura.com

Han Qin
Data & Technology
Ankura
Washington DC, USA
han.qin@ankura.com

Shi Ye
Data & Technology
Ankura
Washington DC, USA
shi.ye@ankura.com

Amy Tsang
Data & Technology
Ankura
New York, NY, USA
amy.tsang@ankura.com

*Abstract*— **Though technology assisted review in electronic discovery has been focusing on text data, the need of advanced analytics to facilitate reviewing multimedia content is on the rise. In this paper, we present several applications of deep learning in computer vision to Technology Assisted Review of image data in legal industry. These applications include image classification, image clustering, and object detection. We use transfer learning techniques to leverage established pretrained models for feature extraction and fine tuning. These applications are first of their kind in the legal industry for image document review. We demonstrate effectiveness of these applications with solving real world business challenges.**

*Keywords- predictive coding, legal document review, deep learning, image classification, image clustering, object detection*

## I. INTRODUCTION

In the legal industry, to reduce cost and time spent in electronic discovery, it has become a common practice to use machine learning for document review, known as predictive coding or TAR (technology assisted review). Both traditional machine learning methods and deep learning algorithms have been studied for tasks such as text classification and document clustering. [2, 3]. However, these applications only deal with text documents. Images, sounds, videos files in the document depository are usually left out for eyes-on review or unreviewed, which would have cost implications if these types of documents made up more than a small fraction of the total review set [1]. Authors in [4] are the first to explore the potentials of TAR for images, using deep learning/transfer learning in image clustering and image classification. However, the examples in [4] are hypothetical ones, instead of being based on real business scenarios.

In this paper, we present several business cases arise from real legal document review projects for applications of image analytics. Specifically, we will present applications of image classification, image clustering, and object detection using transfer learning. The applications of image classification and image clustering are based on pretrained VGG16 model [6], while the object detection uses a Faster R-CNN model [8].

Computer vision with deep learning has made great advances in the last decade. State-of-the-art neural network models have been created using vast amount of training data and computing resources, yielding high accuracies in challenging tasks including image classification and object detection. At the same time, transfer learning techniques have been proven to be effective in applying those techniques to tasks in a wide range of domains.

As a machine learning method, transfer learning is to reuse a model trained for one task to train a model for a second related task [5], which allows faster training with smaller training dataset for the second task using knowledge learnt from the first.

To reuse a pretrained model of a deep neural network with transfer learning, one can either fine tune the model by replacing the last few layers of the deep neural network with new layers and retrain the new model while freezing the weights of the earlier layers of the original model, or extract outputs from a given layer of the pretrained model as features and feed the features to another machine learning algorithm for further training. We use the former approach for image classification and object detection and the latter for image clustering.

In the following sections we will present, for each of the applications, the business cases, the pretrained models used, and the transfer learning approaches. We will also demonstrate the effectiveness of the applications using results on test datasets.

## II. IMAGE CLASSIFICATION

With rapid growth of images on mobile devices, taken by phone cameras or downloaded from social media apps, automatic classification of images will help lawyers filter out images irrelevant to the cases. A simple example is to build a classification model distinguish images of text documents from other pictures. One potential application is to identify images of document from all pictures of a mobile device.

We implemented an image classification application with functions of model training, model validation, and image classifications. It leverages pretrained model VGG16, created by Oxford's Visual Geometry Group [6]. VGG16 is a 16-layer convolutional neural network consisting of 13 convolution layers and three fully connected layers. It was trained on the ImageNet, for classifications of 1000 classes.

For our transfer learning model, we keep the first 13 convolution layers of VGG16 with pretrained weights but replace the three fully connected layers with a single fully connected layer, followed by a binary classifier output layer. The new model uses a smaller size of input images at 150x150 (compared to VGG16's input image size of 240x240). The number of features extracted from the 13 VGG16 convolution layers is 8192 and the 8192 extracted features are the input to fully connected layer. The structure of the fine tune layers is shown in Table 1.

Depending on user's specific image classification needs, the transfer model can be trained with user provided training data to learn the new top layer weights. The application integrates with a document review platform so that a user can select labeled images in a project within the platform for training. However, in case there is not enough training samples available from the project at the training time, it also allows user to supplement the training set by uploading additional labeled images from other sources. In the example of classification of image of documents, the training samples are supplemented with images downloaded from Google image search, consisting of images of text documents as positive samples and images such as landscape, people, and animals as negative samples, total at less than 500 images.

The trained model yields high accuracy. For a test dataset of 2000 images with a 50/50 split between positive and negative samples, the accuracy rate is above 98%. The high accuracy is probably due to VGG16's ability to capture the essential features that distinguish document images from other type images. In fact, as we shall see in the next section, the task of identifying images of documents can be even achieved using image clustering, using features extracted from VGG16.

*Table 1. Fine Tune Layers*

| Layer (type)       | Output Shape  | Param # |
|--------------------|---------------|---------|
| flatten_1 (Flatten)| (None, 8192)  | 0       |
| dense_1 (Dense)    | (None, 256)   | 2097408 |
| dense_2 (Dense)    | (None, 1)     | 257     |

Identifying images of documents is only one example of image classification applications in legal document review. As more business cases of such applications are developed, we plan to build a library of image classification models such that these models can be reused for legal document review engagements.

## III. IMAGE CLUSTERING

The features that VGG16 extracted from a given layer can be used as inputs not only for supervised learning, but also for unsupervised learning tasks such as clustering. Clustering of images allows users to view the images in groups, to explore the categories of the images, and to distinguish relevant images from irrelevant ones.

The image clustering tool is designed as follows: we resize the images to 240x240, as used in VGG16, run them through the VGG16 model and get the outputs from the second fully connected layer ("fc2", the last layer before the classification layer) as extracted features and the number of extracted features is 4096. We then take these features as inputs to a K-means algorithm. The K-means model uses the default parameters as specified in the scikit-learn package.

Image clustering has been used in several document review projects. In one project, we ran clustering on a set of several thousands of images with varying number of clusters and found clusters of non-responsive images so that they can be excluded for review. For example, in another project, we applied image clustering to more than one hundred thousand images and found a cluster which includes only images of human faces.

Another interesting example of image clustering applications is that when it was applied to a set of images with images of documents, all images of documents fall into one cluster. Image clustering could be used to help identify labeled images for training supervised image classification models.

## IV. OBJECT DETECTION

In one of our client engagements, the client requested us to identify all PDF documents with handwritings. There are a few options for the problem -- one can use an NLP API service provided by many commercial AI platforms, or build one's own model using transfer learning. In [7], a handwriting detection model was trained using transfer learning, but it relies on a commercial platform for the training. For our purpose, we chose to develop a standalone TensorFlow application for handwriting detection using transfer learning with fine tune training.

There have been many state-of-the-art algorithms developed to locate objects of interest within an image. Fast R-CNN, Faster R-CNN, YOLO are among the popular ones. For our application, we used the Fast R-CNN model, faster_rcnn_resnet50_coco, trained by Microsoft on the COCO dataset [8], along with Google's Tensorflow Object Detection API for the handwriting detection task. The API provides fine tune algorithm with configurable parameters such as pretrained model choice, image labeling for training purpose, and an inference. With the API, all we need to do is to prepare a training dataset of document images with handwriting, locate handwritings in an image, and train the handwriting recognition model.

For training data, we downloaded 235 google images of documents with handwritings, including handwriting notes, signatures, and hand filled forms. We then used a tool to label the handwriting regions, which are rectangular boxes surrounding the region with class names. The labels were saved in an XML file in PascalVOC format for each of the images. We then converted the xml files into a single csv file and then split it into training and testing set. Finally, the training and testing data were converted into TFRecords to feed into the learning algorithm.

The Object Detection API in Tensorflow provides configuration file, faster_rcnn_resnet50_coco.config, in which we specified the pretrained object detection model name, training and testing input path, class label mapping path, and convolution hyperparameters, and training steps. By choosing most of the default settings together with some customized parameters, the model then was trained. Figure 1 shows the result of handwriting detection from one image.

We converted each PDF document into images page by page. Our goal is to identify those PDF documents that contains handwritings. Therefore, the problem becomes a binary classification problem. A document is considered positive if there is at least one page that contains a handwriting. We assign a probability score to each document. The handwriting model we trained generates zero or more probability scores for each page, each of the scores corresponds to a candidate for a handwriting. The probability score of a document is the maximum score of the all scores from all pages of the document.

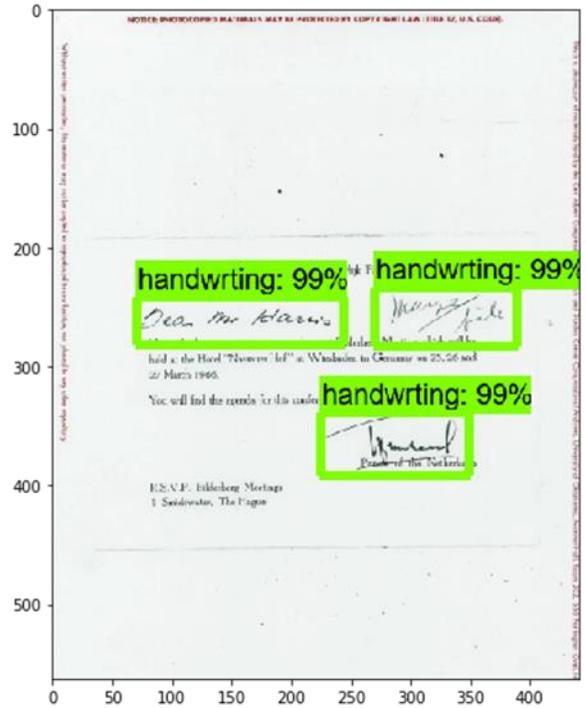
*Figure 1. handwriting detection*

We tested the handwriting detection model on a test dataset sampled from a project. There are 1113 PDF documents with total 4985 pages. Among those documents we manually tagged 291 PDF documents with 655 pages. At document level, using the scoring method described above, the precision recall curve is shown in Figure 2. Performance metrics with four different cut-off scores is shown in Table 2.

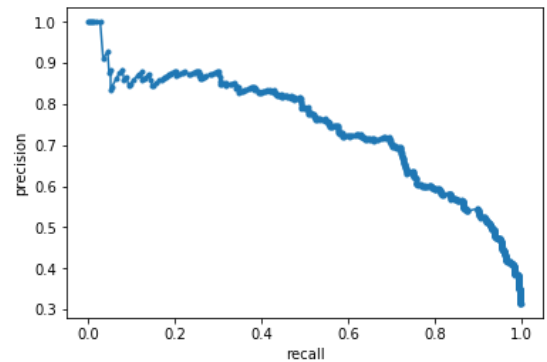
*Figure 2. Precision Recall / Curves for PDF documents*

*Table 2: Performance metrics for selected thresholds*

| Cut-off score | *0.1* | *0.5* | *0.9* | *0.99* |
|---|---|---|---|---|
| Precision | 0.387 | 0.443 | 0.449 | 0.736 |
| Recall | 0.973 | 0.931 | 0.883 | 0.691 |
| F1 | 0.554 | 0.591 | 0.683 | 0.713 |
| Accuracy | 0.59 | 0.663 | 0.738 | 0.845 |

## V. CONCLUSION

In this paper, we presented business cases and technical implementations for deep learning applications in image classification, image clustering, object detection. While these application scenarios are from our document review engagements, we believe that they could become generic applications of image processing using deep learning technology in legal document review. We also believe that there will be more and more applications of image processing technologies in legal document review.

## REFERENCES

[1] *Nicholas M. Pace & Laura Zakaras, "Where the Money Goes: Understanding Litigant Expenditures for Producing Electronic Discovery," RAND at 17 (2012).*

[2] *Chhatwal, R., Huber-Fliflet, N., Keeling, R., Zhang, J. and Zhao, H. (2016). Empirical Evaluations of Preprocessing Parameters' Impact on Predictive Coding's Effectiveness. In Proceedings of 2016 IEEE International Big Data Conference*

[3] *Wei, F., Han, Q., Ye, S., Zhao, H. "Empirical Study of Deep Learning for Text Classification in Legal Document Review" 2018 IEEE International Big Data Conference*

[4] T. Schoinas and D. Esbati, "Technology Assisted Review of Images using Machine Vision," in *2018 IEEE International Conference on Big Data (Big Data)*, Seattle, WA, USA, 2018, pp. 3310–3316.

[5] *Jason Yosinski, Jeff Clune, Yoshua Bengio, and Hod Lipson. 2014. How transferable are features in deep neural networks? In Advances in neural information processing systems. pages 3320–3328.*

[6] *K. Simonyan, A. Zisserman, "Vereep convolutional networks for large-scale image recognition", arXiv:1409.1556, 2014.Y.*

[7] *"Making Sense of Handwritten Sections in Scanned Documents Using the Azure ML Package for Computer Vision and Azure Cognitive Services." Developer Blog (blog), May 7, 2018. https://www.microsoft.com/developerblog/2018/05/07/handwriting-detection-and-recognition-in-scanned-documents-using-azure-ml-package-computer-vision-azure-cognitive-services-ocr/.*

[8] *Ren, Shaoqing, Kaiming He, Ross Girshick, and Jian Sun. "Faster R-CNN: Towards Real-Time Object Detection with Region Proposal Networks." ArXiv:1506.01497 [Cs], June 4, 2015. http://arxiv.org/abs/1506.01497.*